\documentclass[runningheads]{llncs}

\usepackage{eccv}

\usepackage{eccvabbrv}

\usepackage{graphicx}
\usepackage{booktabs}

\usepackage[accsupp]{axessibility}  

\usepackage[pagebackref,breaklinks,colorlinks,citecolor=eccvblue]{hyperref}

\usepackage{orcidlink}

\begin{document}

\title{Attend-Fusion: Efficient Audio-Visual Fusion for Video Classification} 

\titlerunning{Attend-Fusion}

\author{Mahrukh Awan\thanks{These authors contributed equally to this work.}\inst{1} \and
Asmar Nadeem\inst{*}\inst{1} \and \\
Muhammad Junaid Awan\inst{1} \and 
Armin Mustafa\inst{1} \and
Syed Sameed Husain\inst{1}}

\authorrunning{Mahrukh, A., Asmar, N., et al.}

\institute{Centre for Vision, Speech and Signal Processing (CVSSP), University of Surrey, UK\\
\email{\{mahrukh.awan,asmar.nadeem,m.awan,armin.mustafa,sameed.husain\}@surrey.ac.uk}}

\maketitle

\begin{abstract}
Exploiting both audio and visual modalities for video classification is a challenging task, as the existing methods require large model architectures, leading to high computational complexity and resource requirements. Smaller architectures, on the other hand, struggle to achieve optimal performance. In this paper, we propose Attend-Fusion, an audio-visual (AV) fusion approach that introduces a compact model architecture specifically designed to capture intricate audio-visual relationships in video data. Through extensive experiments on the challenging YouTube-8M dataset, we demonstrate that Attend-Fusion achieves an F1 score of 75.64\% with only 72M parameters, which is comparable to the performance of larger baseline models such as Fully-Connected Late Fusion (75.96\% F1 score, 341M parameters). Attend-Fusion achieves similar performance to the larger baseline model while reducing the model size by nearly 80\%, highlighting its efficiency in terms of model complexity. Our work demonstrates that the Attend-Fusion model effectively combines audio and visual information for video classification, achieving competitive performance with significantly reduced model size. This approach opens new possibilities for deploying high-performance video understanding systems in resource-constrained environments across various applications.
\end{abstract}

\section{Introduction}
\label{sec:intro}

Audio-Visual video understanding~\cite{wei2022learning} represents the frontier of video classification, building upon the foundations laid by static image recognition and extending into the complex realm of temporal and multimodal data processing~\cite{iashin2020multi,nadeem2023sem, song2022multimodal, nadeem2024cad, nadeem2024narrativebridge, ding2020self, sun2022human, lv2021progressive, wang2021semantic, shah2022audio}. While datasets like ImageNet~\cite{deng2009imagenet} revolutionized image classification, the emergence of large-scale video datasets such as YouTube-8M~\cite{abu2016youtube} has shifted the focus to video analysis, enabling the evaluation of models' capabilities in multi-label video classification tasks~\cite{lee20182nd,gkalelis2019subclass,li2022multi} and their ability to interpret temporal and multimodal information across visual and audio modalities~\cite{tschannen2020self}.
\begin{figure}[ht]
\centering
\includegraphics[width=1.0\textwidth]{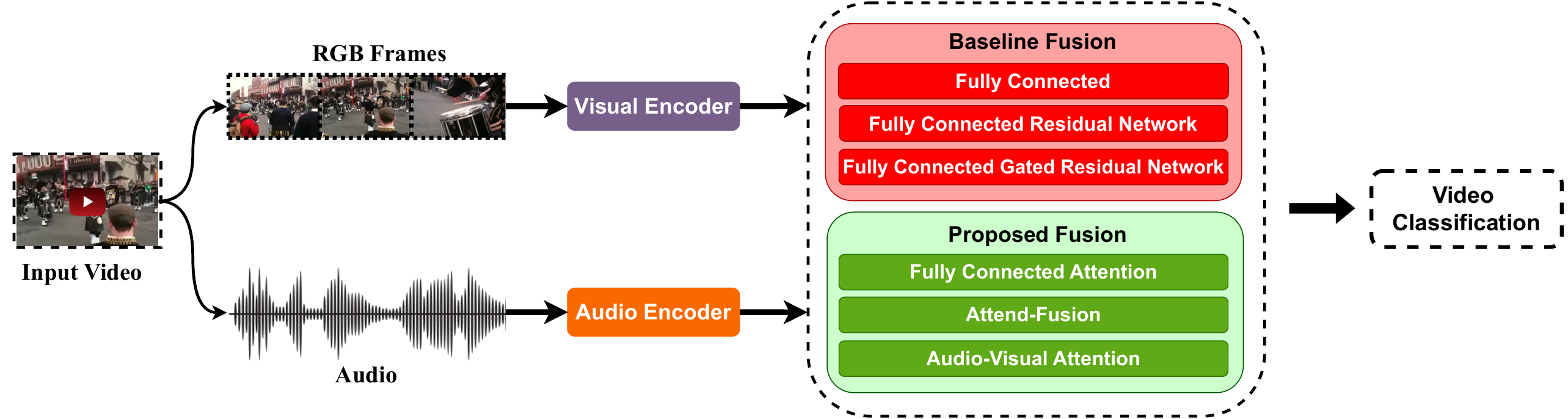}
\caption{ Overview of our proposed audio-visual video classification framework on YouTube-8M dataset, illustrating different fusion mechanisms of audio and visual modalities~\cite{bober2017cultivating,ong2018deep,he2016deep,miech2017learnable}.\vspace{-6mm}}
\label{fig:intro}
\end{figure}

The YouTube-8M dataset~\cite{abu2016youtube} is a large-scale dataset that comprises millions of YouTube videos, each annotated with labels from a diverse vocabulary of 4,716 visual entities~\cite{abu2016youtube}. The task involves predicting these multiple labels for each video, making it a multi-label classification problem. Figure~\ref{fig:intro} illustrates the pipeline of the fusion of visual and audio features for video classification. YouTube-8M is a multimodal dataset featuring visual and audio features, enabling the development of models that integrate both modalities for comprehensive video content analysis~\cite{gkalelis2019subclass,li2022multi}.

Historically, audio~\cite{reddy1976speech} and visual~\cite{poppe2010survey} data have been studied independently, each with its own set of applications, capturing only partial information about the subject of interest, limiting their performance and making them susceptible to noise within that single modality~\cite{wei2022learning}. Recent methods~\cite{planamente2022domain, lv2021progressive, nadeem2024cad} have begun to explore the integration of multiple modalities in audio-visual tasks. Our approach fuses audio and visual modalities (Figure~\ref{fig:intro}) for comprehensive video content understanding. Audio-visual video classification\cite{abu2016youtube} requires effective multimodal fusion, overcoming single-modality limitations. Despite progress, developing efficient fusion methods while maintaining high accuracy remains a challenge. Our research proposes an efficient audio-visual fusion approach for the YouTube-8M dataset, addressing two main questions: (1) How to leverage complementary multimodal information for improved video classification effectively? (2) Can a compact architecture achieve comparable performance to larger models while being more efficient?

Figure~\ref{fig:intro} presents an overview of our proposed audio-visual video classification framework. The input video is first separated into its audio and visual components, which are then processed by their respective encoders to extract meaningful features. These features are fused using different strategies, including baseline methods and our proposed approach. The baseline methods~\cite{bober2017cultivating,ong2018deep,he2016deep,miech2017learnable} consist of fully connected networks, fully connected residual networks, and fully connected gated residual networks, which are commonly used in multimodal learning. On the other hand, our proposed approach introduces three novel architectures: fully connected attention, audio-visual attention, and Attend-Fusion.

Among our proposed methods, Attend-Fusion stands out as the most advanced and effective architecture. It leverages late fusion~\cite{nadeem2024cad} to integrate information from audio and visual modalities effectively. Attend-Fusion enables the model to learn both modality-specific and cross-modal representations. It incorporates attention mechanisms to capture complex temporal and modal relationships in the video data.
To demonstrate the efficacy of Attend-Fusion, we conduct extensive experiments and compare its performance against the baseline methods. In addition to the commonly used Global Average Precision (GAP), we focus on the F1 score, which provides a more comprehensive evaluation of classification performance, especially in multi-label scenarios~\cite{powers2020evaluation, li2022multi, ding2020self, sun2022human}. 

Attend-Fusion employs approximately 72 million parameters, which is significantly smaller compared to leading state-of-the-art models that typically use around 341 million parameters. This represents a reduction of over 80\% in model size, highlighting the efficiency of our approach. This efficiency not only demonstrates the robustness of our methodology but also addresses critical concerns in the field regarding model complexity and computational resources~\cite{nadeem2024cad, song2022multimodal}. By achieving high performance with more compact models, our work contributes to the ongoing efforts to develop more sustainable and deployable AI solutions for video understanding tasks~\cite{planamente2022domain, lv2021progressive}. This has significant implications for real-world applications, as our approach can enable the deployment of accurate video classification systems on resource-constrained devices and facilitate more efficient processing of large-scale video datasets.

Our approach focuses on the collaborative fusion~\cite{wei2022learning} of audio and visual learning components, employing diverse strategies tailored for each modality. As highlighted in Figure~\ref{fig:intro}, we incorporate fully connected networks~\cite{ong2018deep} and various attention mechanisms\cite{hao2022attention, lahat2015multimodal, praveen2022joint} to enhance the learning process from both audio and visual data. We explore a range of fusion methods, from early to late fusion, leveraging the capabilities of these sophisticated techniques. We intend to make videos more comprehensible by strategically exploiting the fusion between audio and visual cues.
Our main contributions in this work are as follows:
\begin{enumerate}
    \item We propose a novel, efficient approach for fusing audio and visual information that achieves high classification accuracy on the YouTube-8M dataset while significantly reducing model size compared to state-of-the-art methods.
    \item We introduce and evaluate advanced attention mechanisms tailored for capturing complex temporal and modal relationships in video data, demonstrating their effectiveness in audio-visual learning.
    \item We provide a comprehensive analysis of the trade-offs between model size, computational efficiency, and classification accuracy, contributing insights for sustainable AI development in video understanding.
\end{enumerate}

Through this research, we aim to bridge the gap between academic advancements and industry applications in the rapidly evolving field of audio-visual machine learning. 

\section{Related Work}
\label{sec: Related Work}
 
Proliferation of multimedia content, advancements in deep learning and high demands for models to understand and interpret AV(audio-visual) data in various tasks have led to recent development in the Deep AV learning domain \cite{zhu2020deep, berghiforecasterflexobm}. Leveraging the complementary nature of audio and visual modalities and learning their shared representation is fundamental to improving the accuracy of semantic understanding in AV learning \cite{wei2022learning}.  \cite{vilaça2022recent} provided an overview of key architectures, methodologies and systematic comparison for deep AV  correlation learning approaches to leverage the complementary information given the complex and dynamic nature of both modalities. Attention Mechanism has proven to be able to curtail noise in audio and visual modalities that complicates the video understanding task \cite{hao2022attention} by selectively focusing on the most relevant parts and effective integration \cite{lahat2015multimodal,praveen2022joint} improving the interpretability of the model \cite{lin2020audiovisual}. 

\subsection{Deep AV Learning}
Deep learning techniques have been extensively employed in audio-visual learning tasks. Convolutional Neural Networks (CNNs) \cite{okazaki2021multi, zhang2017learning, sharafi2023audio} have been widely used for learning spatial features from visual data, while Recurrent Neural Networks (RNNs) \cite{abdullah2020facial} and Long Short-Term Memory (LSTM) networks \cite{bhaskar2023lstm} have been utilized to capture temporal dependencies in sequential data. Recently, transformer networks \cite{vaswani2017attention} have gained popularity in audio-visual learning due to their ability to capture long-range dependencies and model multi-modal interactions effectively. Various transformer-based architectures, such as the Audio-Visual Transformer \cite{nagrani2021attention}, Cross-Modal Transformer \cite{tsai2019multimodal}, and Multi-modal Transformer \cite{gabeur2020multi}, have been proposed for tasks such as audio-visual speech recognition, audio-visual event localization, and multi-modal sentiment analysis. These techniques have been applied to various benchmarked datasets, such as UCF101 \cite{soomro2012ucf101} and YouTube-8M \cite{abuelhaija2016youtube8m}, to address the unique challenges posed by video data, including high dimensionality and temporal dependencies \cite{rehman2021deep}.

\subsection{Video Classification on YouTube-8M Dataset}
The YouTube-8M dataset \cite{abu2016youtube} has become a popular benchmark for video classification tasks, particularly in the context of audio-visual learning. This large-scale dataset consists of millions of YouTube videos annotated with a diverse set of labels, making it a challenging testbed for multi-label video classification. Various approaches have been proposed to tackle the unique challenges posed by this dataset, such as its scale, diversity, and the presence of noisy labels.
Lee et al. \cite{lee20182nd} introduced a collaborative experts model that utilizes a mixture of experts and a classifier to handle the multi-label classification problem on the YouTube-8M dataset. Gkalelis et al. \cite{gkalelis2019subclass} proposed a subclass deep neural network approach that learns a set of subclasses for each label, capturing the complex relationships between video content and labels. Li et al. \cite{li2022multi} presented a multi-modal fusion framework that leverages both audio and visual features to improve video classification performance on the YouTube-8M dataset. Their approach incorporates a cross-modal attention mechanism to selectively focus on relevant audio and visual cues.

In addition to these methods, several studies have explored the use of temporal models, such as LSTMs \cite{bhaskar2023lstm} and attention-based mechanisms \cite{tschannen2020self}, to capture the temporal dependencies and multi-modal interactions in videos from the YouTube-8M dataset. These approaches have demonstrated the importance of effectively leveraging both audio and visual modalities for accurate video classification on this challenging benchmark.

\subsection{Fusion Techniques in AV Learning}
Integrating and synchronizing audio-visual modalities is a fundamental challenge in audio-visual learning \cite{gao2020survey}. Traditional fusion approaches, such as early and late fusion, are still widely used due to their simplicity \cite{lahat2015multimodal, boulahia2021early, sundar2022multimodal}. However, state-of-the-art frameworks often incorporate these approaches with advanced techniques. The Slow Fusion Network \cite{Joze_2020_CVPR} introduces a Multi-modal Transfer Module (MMTM) for feature modality fusion, while the Attention-based Multi-modal Fusion Module (AMFM) \cite{hori2017attention, gu2018hybrid, praveen2022joint, zhang2023facial} incorporates attention mechanisms to selectively fuse audio and visual features.

\subsection{Attention Mechanisms in Audio-Visual Learning}
Attention mechanisms have proven effective in addressing the challenges of audio-visual learning by focusing on the most relevant parts of the input and enabling effective cross-modal integration \cite{hao2022attention, lahat2015multimodal, praveen2022joint}. Several state-of-the-art frameworks have been proposed that incorporate attention mechanisms to capture complex temporal and modal relationships in video data, such as Hierarchical Audiovisual Synchronization \cite{khosravan2019attention}, Generalized Zero-shot Learning with Cross-modal Attention \cite{mercea2022audio, mercea2022temporal}, and Multi-level Attention Fusion Network \cite{cheng2020look, yu2022mm, zhang2023variational}.
Attention mechanisms have also been widely employed in various audio-visual tasks, including event recognition using multi-level attention fusion networks \cite{brousmiche2021multi}, sound localization using instance attention \cite{lin2020audiovisual} and dual attention matching \cite{Wu_2019_ICCV}, and regression-based tasks using recursive joint attention \cite{praveen2023recursive}.

\subsection{Audio-Visual Representation Learning}
Learning effective representations from audio-visual data is crucial for various downstream tasks. Self-supervised learning approaches have been proposed to leverage the inherent synchronization and correspondence between audio and visual modalities \cite{arandjelovic2017look, korbar2018cooperative, morgado2020learning}. These methods aim to learn rich audio-visual representations without the need for explicit human annotations. Contrastive learning techniques have also been employed to learn discriminative audio-visual embeddings \cite{morgado2021audio, chen2021distilling}.

Our proposed approach builds upon the existing methods discussed in this section. We leverage the power of attention mechanisms to effectively integrate audio and visual features while maintaining a compact model architecture. Our approach differs from previous works by introducing a novel combination of early and late fusion strategies, along with attention mechanisms specifically designed for capturing temporal and modal relationships in video data. By incorporating these advanced techniques, we aim to push the boundaries of audio-visual learning and contribute to the state-of-the-art in this rapidly evolving field.

\section{Methodology}

\subsection{Baseline Models}
\label{ssec:baseline_models}
\begin{figure}[t]
    \centering
    \begin{subfigure}{\textwidth}
        \centering
        \includegraphics[width=\linewidth]{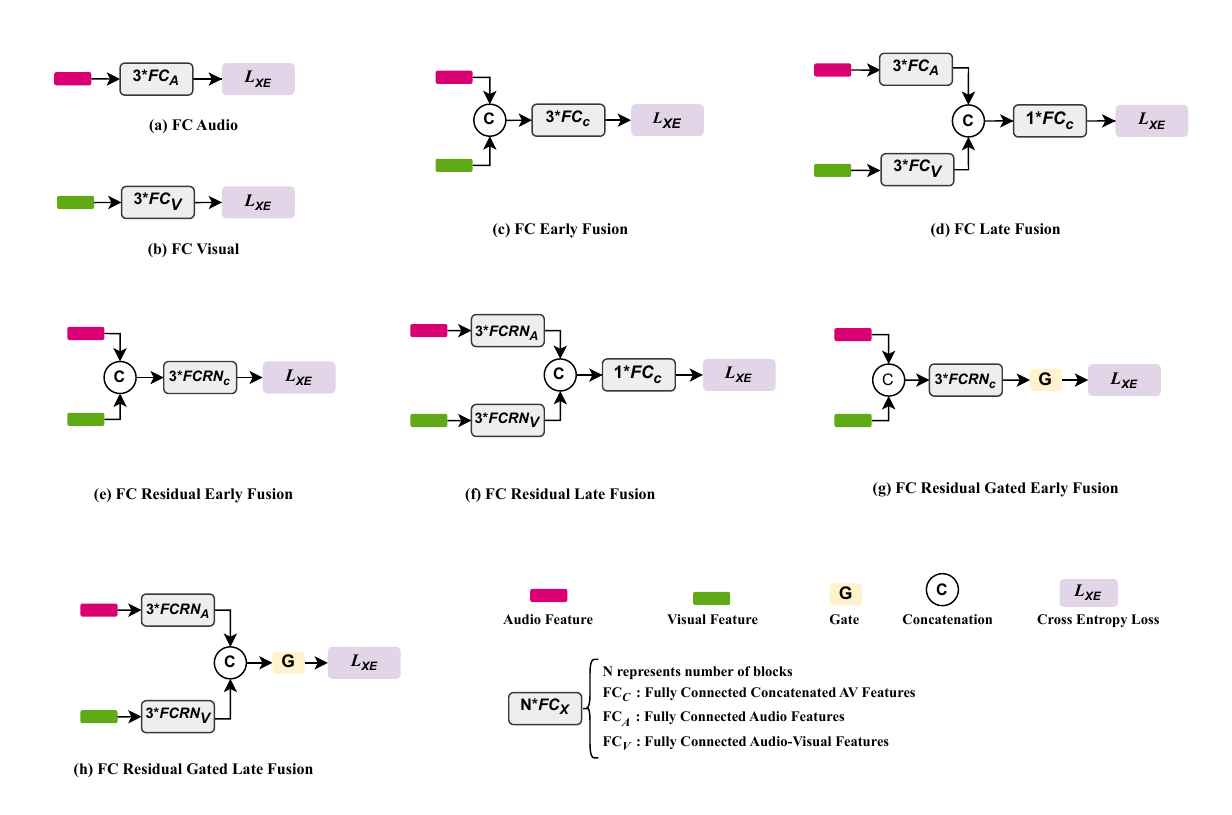}
    \end{subfigure}
    \caption{Illustration of diverse baseline network architectures for audio-visual (AV) fusion: (a,b) Fully Connected (FC) Audio-only and Video-only networks to assess the impact of individual modalities; (c,d) Fully Connected Neural Networks (FCNNs) with early and late fusion strategies; (e,f) Fully Connected Residual Networks (FCRNs) with early and late fusion; and (g,h) Fully Connected Residual Gated Networks (FCRGNs) with early and late fusion, incorporating gating mechanisms for selective feature attention~\cite{bober2017cultivating,ong2018deep,he2016deep,miech2017learnable}. \vspace{-6mm}}
    \label{fig:baseline_models}
\end{figure}

To establish a robust baseline for our study on the YouTube-8M dataset, we implement a series of foundational models derived from prior research in large-scale video classification \cite{bober2017cultivating,ong2018deep,simonyan2014two,hershey2017cnn,karpathy2014large,he2016deep,miech2017learnable,wang2017monkeytyping,chen2017deep}. These models incorporate both unimodal and multimodal approaches with various fusion techniques, as illustrated in Figure \ref{fig:baseline_models}. Our comparative analysis encompasses a range of architectures, including:
\begin{itemize}
\item Fully Connected (FC) Audio and FC Visual networks for unimodal analysis - Figure~\ref{fig:baseline_models}\textcolor{red}{(a,b)} ~\cite{bober2017cultivating}
\item Standard FC with early and late fusion strategies for multimodal analysis - Figure~\ref{fig:baseline_models}\textcolor{red}{(c,d)}~\cite{bober2017cultivating} 
\item FC Residual Networks (FCRN) with early and late fusion - Figure~\ref{fig:baseline_models}\textcolor{red}{(e,f)}~\cite{ong2018deep,he2016deep}
\item FC Residual Gated Networks (FCRGN) with early and late fusion - Figure~\ref{fig:baseline_models}\textcolor{red}{(g,h)}~\cite{ong2018deep,miech2017learnable}
\end{itemize}

\noindent
\textbf{Residual Block}~\cite{ong2018deep,he2016deep}: The residual block used in the baseline models (FCRN and FCRGN) is defined as:
\begin{equation}
\mathbf{y} = \mathcal{F}(\mathbf{x}, {W_i}) + \mathbf{x}
\end{equation}
where $\mathbf{x}$ and $\mathbf{y}$ are the input and output of the block, respectively, $\mathcal{F}$ represents the residual mapping, and ${W_i}$ are the learnable weights.

\noindent
\textbf{Gating Mechanism}~\cite{ong2018deep,miech2017learnable}: The FCRGN models incorporate a gating mechanism to control the flow of information. The gating operation is defined as:
\begin{equation}
\mathbf{g} = \sigma(\mathbf{W}_g \mathbf{x} + \mathbf{b}_g)
\end{equation}
\begin{equation}
\mathbf{y} = \mathbf{g} \odot \mathcal{F}(\mathbf{x}, {W_i}) + (1 - \mathbf{g}) \odot \mathbf{x}
\end{equation}
where $\mathbf{g}$ is the gating vector, $\sigma$ is the sigmoid activation function, $\mathbf{W}_g$ and $\mathbf{b}_g$ are the learnable weights and biases of the gating layer, and $\odot$ denotes element-wise multiplication.\\
\noindent
\textbf{Input Features}: We employ the same input features as used in \cite{ong2018deep}, which are video-level mean and standard deviation features (MF+STD). The input features are extracted from pre-trained Inception networks. Specifically, the frame-level features are obtained from two separate Inception networks, one for video (1024-dimensional) and another for audio (128-dimensional).\\
\noindent
\textbf{Unimodal Networks}: The FC Audio and FC Visual networks consist of three fully connected layers each, assessing the isolated impact of audio and visual modalities, respectively.\\
\noindent
\textbf{Multimodal Networks}: The standard FC network comprises three fully connected layers and is evaluated with both early and late fusion strategies. The FCRN architecture incorporates skip connections to facilitate complex feature learning and implicit regularization. The FCRGN enhances the FCRN by introducing context gating mechanisms for selective feature attention.

\subsection{Proposed Models}
\label{sec:proposed}
While the baseline models employ fully connected layers for processing audio-visual features, our proposed method leverages attention mechanism to dynamically focus on the most relevant parts of the input and capture long-range dependencies. Figure \ref{fig:attention_models} provides an overview of our attention-based network architectures.
\subsubsection{3.2.1 FC Attention Network}
\label{ssec:fc_attention}
\begin{figure*}[t!]

    \begin{subfigure}{\textwidth}
        \centering
        \includegraphics[width=\linewidth]{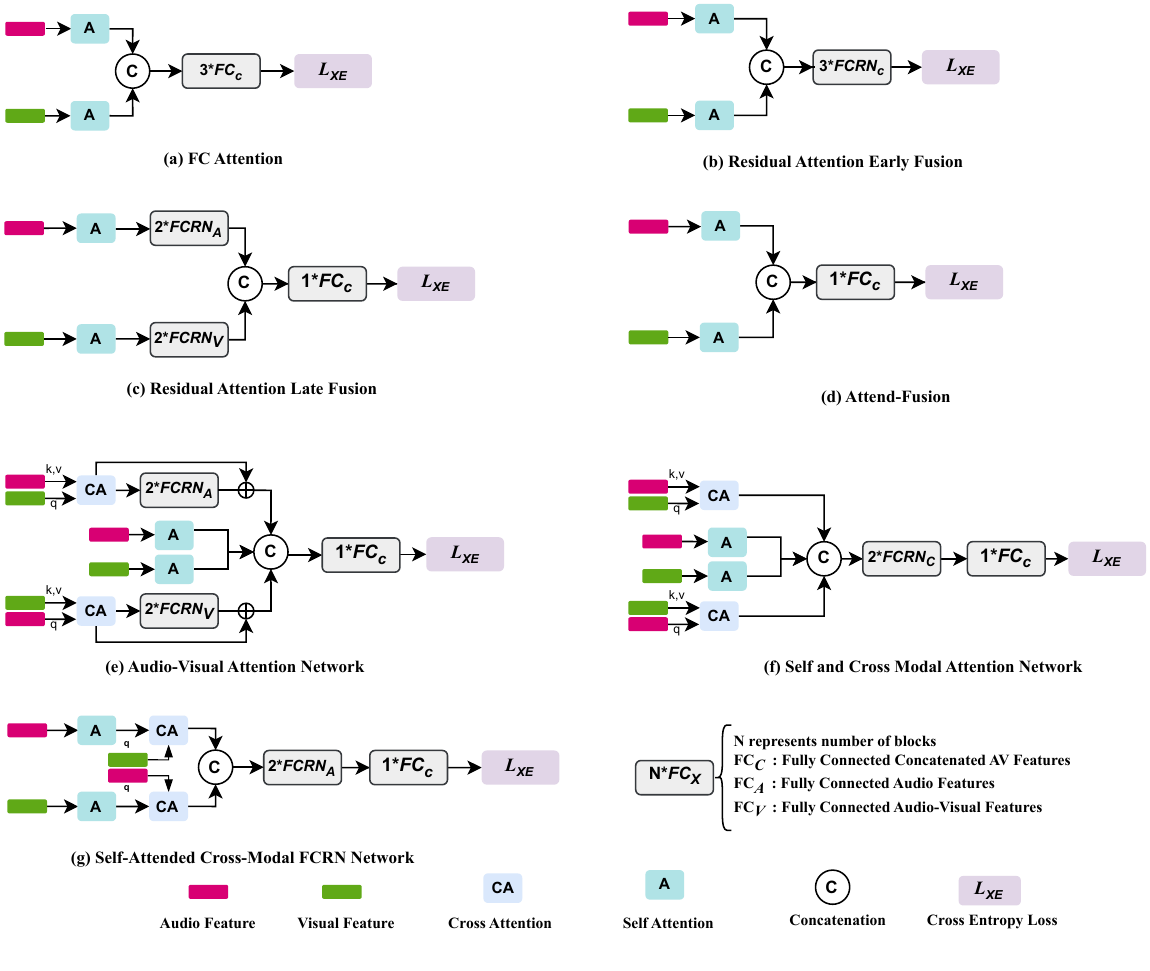}
    \end{subfigure}
   \caption{Attention-based network architectures: (a) FC Attention Network; (b,c) FC Residual Attention Networks with early and late fusion; (d,e) AV Attention Fusion Networks; (f) Self and Cross Modal Attention Network; (g) Network with Self-Attended Features for Cross Modal Attention.\vspace{-6mm}}
   \label{fig:attention_models}

\end{figure*}

The FC Attention Network (Figure~\ref{fig:attention_models}\textcolor{red}{(a)}) integrates self-attention~\cite{vaswani2017attention} mechanisms to prioritize salient features within each modality. The attended audio and visual features are fused through concatenation, followed by fully connected layers with ReLU activation and dropout regularization. The final classification layer employs a sigmoid activation function for multi-label classification.\\
\textbf{Self-Attention Mechanism:} Let $\mathbf{X} \in \mathbb{R}^{N \times d}$ be the input feature matrix, where $N$ is the number of features and $d$ is the feature dimension. The self-attention mechanism computes the attended features $\mathbf{X}_{att}$ as follows:
\begin{equation}
\mathbf{Q} = \mathbf{X}\mathbf{W}_Q, \quad \mathbf{K} = \mathbf{X}\mathbf{W}_K, \quad \mathbf{V} = \mathbf{X}\mathbf{W}_V
\end{equation}
\begin{equation}
\mathbf{X}_{att} = \text{softmax}\left(\frac{\mathbf{Q}\mathbf{K}^T}{\sqrt{d}}\right)\mathbf{V}
\end{equation}
where $\mathbf{W}_Q$, $\mathbf{W}_K$, and $\mathbf{W}_V$ are learnable weight matrices.

\subsubsection{3.2.2 Residual Attention Networks}
The Fully Connected Residual Attention Networks (Figure~\ref{fig:attention_models}\textcolor{red}{(b,c)}) combine attention mechanisms with residual learning \cite{feichtenhofer2017spatiotemporal}. We experiment with early and late fusion variants to understand the impact of multimodal integration timing on the network's performance. The residual block is similar to what we use in Section~\ref{ssec:baseline_models}.

\subsubsection{3.2.3 Attend-Fusion}
Attend-Fusion (Figure~\ref{fig:attention_models}\textcolor{red}{(d)}) is a novel architecture that effectively combines attention mechanisms and multi-stage fusion for audio-visual video classification. It processes audio and visual features separately through attention networks, which consist of fully connected layers and self-attention mechanisms, similar to in Section~\ref{ssec:fc_attention}\textcolor{red}{.1}. The self-attention allows the model to focus on the most relevant features within each modality.
The attended audio and visual features are then fused using a late fusion strategy, where they are concatenated along the feature dimension. The fused features undergo further processing through fully connected layers, which learn to capture complex interactions between the modalities.

Attend-Fusion's key advantages include its ability to learn modality-specific and cross-modal representations at different stages, and its compact and efficient design. By leveraging attention mechanisms and late fusion, Attend-Fusion achieves comparative performance while maintaining a smaller model size compared to baseline approaches.
\subsubsection{3.2.4    Audio-Visual Attention Network}
The Audio-Visual Attention Network (Figure~\ref{fig:attention_models}\textcolor{red}{(e)}) introduces cross-modal attention to capture nuanced interactions between audio and visual streams. The architecture employs a hierarchical attention approach, with self-attention mechanisms followed by cross-modal attention, to learn refined multimodal representations.\\
\textbf{Cross-Modal Attention:} Let $\mathbf{X}a$ and $\mathbf{X}v$ be the attended audio and visual features, respectively. The cross-modal attention mechanism computes the audio-guided visual features $\mathbf{X}{v|a}$ and the visual-guided audio features $\mathbf{X}{a|v}$ as follows:
\begin{equation}
\mathbf{Q}_a = \mathbf{X}a\mathbf{W}{Q_a}, \quad \mathbf{K}_v = \mathbf{X}v\mathbf{W}{K_v}, \quad \mathbf{V}v = \mathbf{X}v\mathbf{W}{V_v}
\end{equation}
\begin{equation}
\mathbf{X}{v|a} = \text{softmax}\left(\frac{\mathbf{Q}_a\mathbf{K}_v^T}{\sqrt{d}}\right)\mathbf{V}_v
\end{equation}
\begin{equation}
\mathbf{Q}_v = \mathbf{X}v\mathbf{W}{Q_v}, \quad \mathbf{K}_a = \mathbf{X}a\mathbf{W}{K_a}, \quad \mathbf{V}a = \mathbf{X}a\mathbf{W}{V_a}
\end{equation}
\begin{equation}
\mathbf{X}{a|v} = \text{softmax}\left(\frac{\mathbf{Q}_v\mathbf{K}_a^T}{\sqrt{d}}\right)\mathbf{V}_a
\end{equation}
The attended features $\mathbf{X}{v|a}$ and $\mathbf{X}{a|v}$ are then concatenated with the self-attended features $\mathbf{X}_a$ and $\mathbf{X}_v$ for further processing.

\subsubsection{3.2.5 Self and Cross Modal Attention Network}
The Self and Cross Modal Attention Network (Figure~\ref{fig:attention_models}\textcolor{red}{(f)}) extends the Audio-Visual Attention Network by incorporating additional self-attention layers after the cross-modal attention. This allows the network to refine the learned representations further by capturing intra-modal dependencies.
\subsubsection{3.2.6 Self-Attended Cross-Modal FCRN Network}
The Self-Attended Cross-Modal FCRN Network (Figure~\ref{fig:attention_models}\textcolor{red}{(g)}) combines the self-attention and cross-modal attention mechanisms with the residual learning framework. The network employs residual blocks to facilitate the learning of complex feature interactions while leveraging the attention mechanisms to focus on relevant information.

\subsection{Loss Function}
We employ the cross-entropy loss function for multi-label classification:
\begin{equation}
\mathcal{L} = -\frac{1}{N}\sum_{i=1}^N\sum_{c=1}^C y_{i,c}\log(\hat{y}_{i,c}) + (1 - {y}_{i,c})\log(1 - \hat{y}_{i,c})
\end{equation}
where $N$ is the number of samples, $C$ is the number of classes, $y_{i,c}$ is the ground truth label for sample $i$ and class $c$, and $\hat{y}_{i,c}$ is the predicted probability.

\section{Experimentation and Results}

\subsection{Implementation Details}
All models are implemented using PyTorch and trained on NVIDIA RTX3090 GPUs. We apply dropout \cite{srivastava2014dropout} with a rate of 0.4 to the fully connected layers to prevent overfitting. The dimensions of fully connected layers are set to 8K for the baseline models and 2K for the proposed models, and the dimensions of the attention layers are set to 1024. All models are trained using the AdamW optimizer \cite{loshchilov2017decoupled} with a learning rate of 0.0001. The models are trained for 20 epochs on the YouTube-8M dataset. 

\subsection{Evaluation Metrics}
We evaluate the performance of our models using the Global Average Precision (GAP) metric and F1 score. The GAP metric is the mean Average Precision (AP) across all classes. The AP for a single class is defined as:
\begin{equation}
AP = \sum_{k=1}^n P(k)\Delta r(k)
\end{equation}
where $n$ is the number of test samples, $P(k)$ is the precision at cut-off $k$ in the list, and $\Delta r(k)$ is the change in recall from items $k-1$ to $k$.\\

The F1 score is the harmonic mean of precision and recall, providing a balanced measure of the model's performance. It is calculated as follows:
\begin{equation}
F1 = 2 \cdot \frac{precision \cdot recall}{precision + recall}
\end{equation}
where precision is the fraction of true positive predictions among all positive predictions, and recall is the fraction of true positive predictions among all actual positive instances.

\subsection{Results and Discussion}

\subsubsection{Overview}

In this section, we present a comprehensive evaluation of our proposed audio-visual video classification models, comparing their performance with baseline approaches on the YouTube-8M dataset. We report both the Global Average Precision (GAP) and F1 scores for each model, providing a holistic view of their classification accuracy. Additionally, we conduct ablation studies to investigate the impact of various components and design choices on the performance of our models.

\subsubsection{Quantitative Results}

Table \ref{tab:results} presents the performance comparison of our proposed models with the baseline models on the YouTube-8M dataset, reporting both GAP and F1 scores. The results demonstrate that the attention-based models consistently outperform the baseline models, with the Self-Attended Cross-Modal FCRN Network achieving the highest GAP of 80.68\%. However, the Attend-Fusion model achieves the F1 score of 75.64\% with significantly fewer parameters (72M) compared to the best-performing baseline model, FC Late Fusion (341M parameters), which achieves an F1 score of 75.96\%.

Among the baseline models, the FC Late Fusion achieves the best performance with a GAP of 80.87\% and an F1 score of 75.96\%. The FC Audio and FC Visual models, which rely on a single modality, perform significantly worse than the multimodal approaches, highlighting the importance of leveraging both audio and visual information for accurate video classification.

\begin{table}[h]
\vspace{-4mm}
\caption{Comparison of Baseline Results, Introduced Models, and Ablation Results}
\label{tab:results}
\begin{tabular}{l@{\hspace{1cm}}c@{\hspace{0.3cm}}c@{\hspace{0.3cm}}c}
\toprule
\textbf{Method} & \textbf{GAP\%} & \textbf{F1 Score\%} & \textbf{Params(M)} \\
\midrule
\textbf{Baseline Models} & & & \\
\midrule
FC Audio~\cite{bober2017cultivating}& 50.32 & 49.09 & 103\\
FC Visual~\cite{bober2017cultivating} & 76.25 & 72.69 & 110\\
FC Early Fusion~\cite{bober2017cultivating} & 80.16 & 75.25 &111 \\
\textbf{FC Late Fusion}~\cite{bober2017cultivating} &\textbf{ 80.87} & \textbf{75.96} & \textbf{341}\\
FC Residual Early Fusion~\cite{ong2018deep,he2016deep} & 80.73& 75.79 & 175\\
FC Residual Late Fusion~\cite{ong2018deep,he2016deep} & 79.10 & 74.37 & 341 \\
FC Residual Gated Early Fusion~\cite{ong2018deep,miech2017learnable} & 79.75 & 74.20 & 175 \\
FC Residual Gated Late Fusion~\cite{ong2018deep,miech2017learnable} & 79.20 & 74.24 & 416 \\
\midrule
\textbf{Our Models} & & & \\
\midrule
FC Attention & 79.76 & 75.09 & 83 \\
Residual Attention Early Fusion &80.53 & 75.34 & 79\\
Residual Attention Late Fusion & 80.59 &75.56 & 83 \\
\textbf{Attend-Fusion} & 80.55 & \textbf{75.64} & \textbf{72}\\
Audio-Visual Attention Network & 80.44 & 75.56 & 113 \\
Self and Cross Modal Attention Network& 80.61 &75.52& 172 \\
Self-Attended Cross-Modal FCRN Network & \textbf{80.68} & 75.45 & 172 \\

\bottomrule
\end{tabular}
\end{table}
\vspace{-0mm}

Our proposed attention-based models demonstrate superior performance compared to baselines. The Self-Attended Cross-Modal FCRN Network achieves the highest GAP of 80.68\%, while the Attend-Fusion model achieves the best F1 score of 75.64\% with a much smaller model size (72M parameters). These results show the effectiveness of attention mechanisms in capturing relevant features and dependencies within and between modalities.

\begin{figure}[t]
\centering
\includegraphics[width=1.0\textwidth]{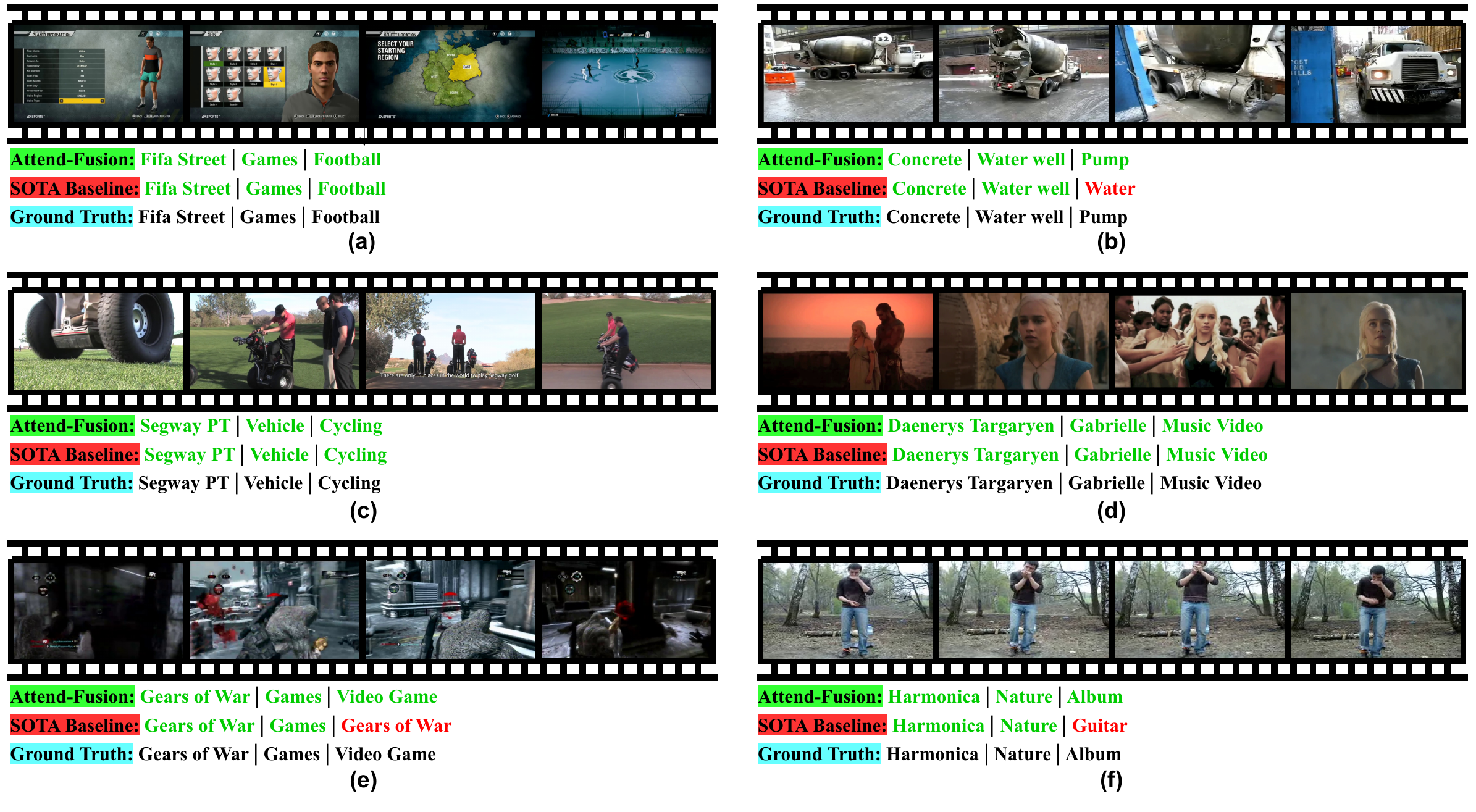}
\caption{Qualitative results comparing the top-3 predictions of our proposed Attend-Fusion model, the state-of-the-art (SOTA) baseline, and the ground truth labels on representative examples from the YouTube-8M dataset.\vspace{-6mm}}
\label{fig:qual_fig}
\end{figure}

\subsubsection{Qualitative Analysis}

To gain further insights into the performance of our proposed Attend-Fusion model, we conduct a qualitative analysis by examining its predictions on a set of representative examples from the YouTube-8M dataset. Figure~\ref{fig:qual_fig} presents a comparison of the top-3 predictions made by our Attend-Fusion model, the state-of-the-art (SOTA) baseline, and the ground-truth labels for six different videos.

In Figure~\ref{fig:qual_fig}(a), both Attend-Fusion and the SOTA baseline correctly predict all three labels: "Fifa Street", "Games", and "Football". This shows their ability to accurately classify sports-related video content.
The Figure~\ref{fig:qual_fig}(b) shows a video related to a water well and pump. Attend-Fusion correctly predicts all three labels, including "Concrete", "Water well", and "Pump". In contrast, the SOTA baseline misses the "Pump" label and predicts "Water" instead, highlighting the superior performance of our model in capturing fine-grained details.

In Figure~\ref{fig:qual_fig}(c), a video is featuring a Segway PT vehicle and cycling. Both Attend-Fusion and the SOTA baseline accurately predict all three labels, showcasing their effectiveness in classifying transportation-related content. In Figure~\ref{fig:qual_fig}(d), both models correctly predict the labels "Daenerys Targaryen", "Gabrielle" and "Music Video", demonstrating their ability to recognize characters and identify the video genre.
Figure~\ref{fig:qual_fig}(e) presents a video from the game "Gears of War". While both models correctly predict the game title and the "Games" label, Attend-Fusion accurately classifies it as a "Video Game", whereas the SOTA baseline predicts "Gears of War" as the third label, which is redundant.

Figure~\ref{fig:qual_fig}(f) shows a video related to a harmonica, nature, and an album. Attend-Fusion correctly predicts all three labels, showcasing its ability to identify musical instruments and themes. The SOTA baseline, however, misclassifies the "Album" label as "Guitar", indicating its limitations in distinguishing between different musical elements.
These qualitative examples demonstrate the superior performance of our Attend-Fusion model in accurately classifying videos across various domains, including sports, transportation, gaming, music, and more. The model's ability to capture fine-grained details and maintain coherence in its predictions highlights the effectiveness of the attention mechanism in integrating audio and visual information for enhanced video classification.

The Attend-Fusion model strikes a balance between performance and efficiency, making it a promising choice for real-world applications. This efficiency is particularly valuable in scenarios where real-time video classification is needed or when deploying models on resource-constrained devices.

\subsection{Ablation Studies}

To further investigate the contributions of different components in our proposed models, we conduct ablation studies as shown in Table \ref{tab:ab_base_results} and Table \ref{tab:ab_our_results}.

Table \ref{tab:ab_base_results} presents the ablation results for the best-performing baseline model, FC Late Fusion. We evaluate the impact of using only the audio or visual modality. The results show that using only the audio modality (Audio Only) leads to a significant drop in performance, with a GAP of 50.32\% and an F1 score of 49.09\%. Similarly, using only the visual modality (Visual Only) also results in a performance decrease, with a GAP of 76.25\% and an F1 score of 72.69\%. These findings highlight the importance of multimodal fusion for achieving high classification accuracy.

\begin{table}[h]
\vspace{-6mm}
\centering
\small
\begin{minipage}{.48\textwidth}
  \centering
  \caption{Ablation Results: Best Baseline.}
  \label{tab:ab_base_results}
  \begin{tabular}{l@{\hspace{0.2cm}}c@{\hspace{0.2cm}}c@{\hspace{0.2cm}}c}
  \toprule
  \textbf{Method} & \textbf{GAP} & \textbf{F1} & \textbf{Params} \\
  & \textbf{(\%)} & \textbf{(\%)} & \textbf{(M)} \\
  \midrule
  FC Late Fusion & \textbf{80.87} & \textbf{75.96} & \textbf{341}\\
  \midrule
  \textbf{Ablation} & & & \\
  Audio Only & 50.32 & 49.09 & 103\\
  Visual Only & 76.25 & 72.69 & 110\\
  \bottomrule
  \end{tabular}
\end{minipage}%
\hfill
\begin{minipage}{.48\textwidth}
  \centering
  \caption{Ablation Results: Best Proposed Model.}
  \label{tab:ab_our_results}
  \begin{tabular}{l@{\hspace{0.2cm}}c@{\hspace{0.2cm}}c@{\hspace{0.2cm}}c}
  \toprule
  \textbf{Method} & \textbf{GAP} & \textbf{F1} & \textbf{Params} \\
  & \textbf{(\%)} & \textbf{(\%)} & \textbf{(M)} \\
  \midrule
  Attend-Fusion & \textbf{80.55} & \textbf{75.64} & \textbf{72}\\
  \midrule
  \textbf{Ablation} & & & \\
  Audio Only & 49.59 & 49.15 & 35\\
  Visual Only & 76.96 & 71.87 & 37 \\
  No Attention & 79.06 & 73.31& 22 \\
  \bottomrule
  \end{tabular}
\end{minipage}
\vspace{-0mm}
\end{table}

Table \ref{tab:ab_our_results} presents the ablation results for our most efficient model, Attend-Fusion. We investigate the impact of removing the attention mechanism (No Attention) and using only the audio or visual modality (Audio Only and Visual Only). Removing the attention mechanism leads to a drop in performance, with a GAP of 79.06\% and an F1 score of 73.31\%. This demonstrates the effectiveness of attention in capturing relevant features and improving classification accuracy. Using only the audio or visual modality results in a significant performance decrease, confirming the importance of multimodal fusion in the Attend-Fusion architecture.

\section{Conclusion}

The experimental results demonstrate the superiority of our proposed attention-based models for audio-visual video classification. By effectively leveraging the complementary information from both modalities and selectively attending to relevant features, our models achieve state-of-the-art performance on the challenging YouTube-8M dataset.

The Attend-Fusion model stands out as a particularly promising approach, achieving competitive performance with a significantly smaller model size compared to the baselines. This efficiency makes it well-suited for real-world applications where computational resources are limited, such as mobile devices or edge computing scenarios. These findings can guide future research in designing more efficient and accurate audio-visual video classification models.

Overall, our work demonstrates the potential of attention-based models for tackling the challenging task of audio-visual video classification. By effectively leveraging the rich information present in both modalities and selectively attending to relevant features, our models push the boundaries of video understanding and pave the way for more advanced and efficient approaches in this field.

\section{Acknowledgement}
This research was partly supported by the British Broadcasting Corporation Research and Development (BBC R\&D), Engineering and Physical Sciences Research Council (EPSRC) Grant EP/V038087/1 “BBC Prosperity Partnership: Future Personalised Object-Based Media Experiences Delivered at Scale Anywhere”.

\bibliographystyle{splncs04}
\bibliography{main}

\begin{thebibliography}{10}
\providecommand{\url}[1]{\texttt{#1}}
\providecommand{\urlprefix}{URL }
\providecommand{\doi}[1]{https://doi.org/#1}

\bibitem{abdullah2020facial}
Abdullah, M., Ahmad, M., Han, D.: Facial expression recognition in videos: An
  cnn-lstm based model for video classification. In: 2020 International
  Conference on Electronics, Information, and Communication (ICEIC). pp.~1--3.
  IEEE (2020)

\bibitem{abu2016youtube}
Abu-El-Haija, S., Kothari, N., Lee, J., Natsev, P., Toderici, G., Varadarajan,
  B., Vijayanarasimhan, S.: Youtube-8m: A large-scale video classification
  benchmark. arXiv preprint arXiv:1609.08675  (2016)

\bibitem{abuelhaija2016youtube8m}
Abu-El-Haija, S., Kothari, N., Lee, J., Natsev, P., Toderici, G., Varadarajan,
  B., Vijayanarasimhan, S.: Youtube-8m: A large-scale video classification
  benchmark (2016)

\bibitem{arandjelovic2017look}
Arandjelovic, R., Zisserman, A.: Look, listen and learn. In: Proceedings of the
  IEEE international conference on computer vision. pp. 609--617 (2017)

\bibitem{berghiforecasterflexobm}
Berghi, D., Cieciura, C., Einabadi, F., Glancy, M., Camilleri, O.C., Foster,
  P., Nadeem, A., Sardari, F., Zhao, J., Volino, M., et~al.: Forecasterflexobm:
  A multi-view audio-visual dataset for flexible object-based media production

\bibitem{bhaskar2023lstm}
Bhaskar, S., Thasleema, T.: Lstm model for visual speech recognition through
  facial expressions. Multimedia Tools and Applications  \textbf{82}(4),
  5455--5472 (2023)

\bibitem{bober2017cultivating}
Bober-Irizar, M., Husain, S., Ong, E.J., Bober, M.: Cultivating dnn diversity
  for large scale video labelling. arXiv preprint arXiv:1707.04272  (2017)

\bibitem{boulahia2021early}
Boulahia, S.Y., Amamra, A., Madi, M.R., Daikh, S.: Early, intermediate and late
  fusion strategies for robust deep learning-based multimodal action
  recognition. Machine Vision and Applications  \textbf{32}(6), ~121 (2021)

\bibitem{brousmiche2021multi}
Brousmiche, M., Rouat, J., Dupont, S.: Multi-level attention fusion network for
  audio-visual event recognition. arXiv preprint arXiv:2106.06736  (2021)

\bibitem{chen2017deep}
Chen, L., Srivastava, S., Duan, Z., Xu, C.: Deep cross-modal audio-visual
  generation. In: Proceedings of the on Thematic Workshops of ACM Multimedia
  2017. pp. 349--357 (2017)

\bibitem{chen2021distilling}
Chen, P., Liu, S., Zhao, H., Jia, J.: Distilling knowledge via knowledge
  review. In: Proceedings of the IEEE/CVF conference on computer vision and
  pattern recognition. pp. 5008--5017 (2021)

\bibitem{cheng2020look}
Cheng, Y., Wang, R., Pan, Z., Feng, R., Zhang, Y.: Look, listen, and attend:
  Co-attention network for self-supervised audio-visual representation
  learning. In: Proceedings of the 28th ACM International Conference on
  Multimedia. pp. 3884--3892 (2020)

\bibitem{deng2009imagenet}
Deng, J., Dong, W., Socher, R., Li, L.J., Li, K., Fei-Fei, L.: Imagenet: A
  large-scale hierarchical image database pp. 248--255 (2009)

\bibitem{ding2020self}
Ding, Y., Xu, Y., Zhang, S.X., Cong, Y., Wang, L.: Self-supervised learning for
  audio-visual speaker diarization. In: ICASSP 2020-2020 IEEE International
  Conference on Acoustics, Speech and Signal Processing (ICASSP). pp.
  4367--4371. IEEE (2020)

\bibitem{feichtenhofer2017spatiotemporal}
Feichtenhofer, C., Pinz, A., Wildes, R.P.: Spatiotemporal multiplier networks
  for video action recognition. In: Proceedings of the IEEE conference on
  computer vision and pattern recognition. pp. 4768--4777 (2017)

\bibitem{gabeur2020multi}
Gabeur, V., Sun, C., Alahari, K., Schmid, C.: Multi-modal transformer for video
  retrieval. In: Computer Vision--ECCV 2020: 16th European Conference, Glasgow,
  UK, August 23--28, 2020, Proceedings, Part IV 16. pp. 214--229. Springer
  (2020)

\bibitem{gao2020survey}
Gao, J., Li, P., Chen, Z., Zhang, J.: A survey on deep learning for multimodal
  data fusion. Neural Computation  \textbf{32}(5),  829--864 (2020)

\bibitem{gkalelis2019subclass}
Gkalelis, N., Mezaris, V.: Subclass deep neural networks: re-enabling neglected
  classes in deep network training for multimedia classification. In:
  International Conference on Multimedia Modeling. pp. 227--238. Springer
  (2019)

\bibitem{gu2018hybrid}
Gu, Y., Yang, K., Fu, S., Chen, S., Li, X., Marsic, I.: Hybrid attention based
  multimodal network for spoken language classification. In: Proceedings of the
  Conference. association for Computational Linguistics. meeting. vol.~2018,
  p.~2379. NIH Public Access (2018)

\bibitem{hao2022attention}
Hao, Y., Wang, S., Cao, P., Gao, X., Xu, T., Wu, J., He, X.: Attention in
  attention: Modeling context correlation for efficient video classification.
  IEEE Transactions on Circuits and Systems for Video Technology
  \textbf{32}(10),  7120--7132 (2022)

\bibitem{he2016deep}
He, K., Zhang, X., Ren, S., Sun, J.: Deep residual learning for image
  recognition. In: Proceedings of the IEEE conference on computer vision and
  pattern recognition. pp. 770--778 (2016)

\bibitem{hershey2017cnn}
Hershey, S., Chaudhuri, S., Ellis, D.P., Gemmeke, J.F., Jansen, A., Moore,
  R.C., Plakal, M., Platt, D., Saurous, R.A., Seybold, B., et~al.: Cnn
  architectures for large-scale audio classification. In: 2017 ieee
  international conference on acoustics, speech and signal processing (icassp).
  pp. 131--135. IEEE (2017)

\bibitem{hori2017attention}
Hori, C., Hori, T., Lee, T.Y., Zhang, Z., Harsham, B., Hershey, J.R., Marks,
  T.K., Sumi, K.: Attention-based multimodal fusion for video description. In:
  Proceedings of the IEEE international conference on computer vision. pp.
  4193--4202 (2017)

\bibitem{iashin2020multi}
Iashin, V., Rahtu, E.: Multi-modal dense video captioning. In: Proceedings of
  the IEEE/CVF Conference on Computer Vision and Pattern Recognition Workshops.
  pp. 958--959 (2020)

\bibitem{Joze_2020_CVPR}
Joze, H.R.V., Shaban, A., Iuzzolino, M.L., Koishida, K.: Mmtm: Multimodal
  transfer module for cnn fusion. In: Proceedings of the IEEE/CVF Conference on
  Computer Vision and Pattern Recognition (CVPR) (June 2020)

\bibitem{karpathy2014large}
Karpathy, A., Toderici, G., Shetty, S., Leung, T., Sukthankar, R., Fei-Fei, L.:
  Large-scale video classification with convolutional neural networks. In:
  Proceedings of the IEEE conference on Computer Vision and Pattern
  Recognition. pp. 1725--1732 (2014)

\bibitem{khosravan2019attention}
Khosravan, N., Ardeshir, S., Puri, R.: On attention modules for audio-visual
  synchronization. In: CVPR Workshops. pp. 25--28 (2019)

\bibitem{korbar2018cooperative}
Korbar, B., Tran, D., Torresani, L.: Cooperative learning of audio and video
  models from self-supervised synchronization. Advances in Neural Information
  Processing Systems  \textbf{31} (2018)

\bibitem{lahat2015multimodal}
Lahat, D., Adali, T., Jutten, C.: Multimodal data fusion: an overview of
  methods, challenges, and prospects. Proceedings of the IEEE  \textbf{103}(9),
   1449--1477 (2015)

\bibitem{lee20182nd}
Lee, J., Reade, W., Sukthankar, R., Toderici, G., et~al.: The 2nd youtube-8m
  large-scale video understanding challenge. In: Proceedings of the European
  Conference on Computer Vision (ECCV) Workshops. pp.~0--0 (2018)

\bibitem{li2022multi}
Li, X., Wu, H., Li, M., Liu, H.: Multi-label video classification via coupling
  attentional multiple instance learning with label relation graph. Pattern
  Recognition Letters  \textbf{156},  53--59 (2022)

\bibitem{lin2020audiovisual}
Lin, Y.B., Wang, Y.C.F.: Audiovisual transformer with instance attention for
  audio-visual event localization. In: Proceedings of the Asian Conference on
  Computer Vision (2020)

\bibitem{loshchilov2017decoupled}
Loshchilov, I., Hutter, F.: Decoupled weight decay regularization. arXiv
  preprint arXiv:1711.05101  (2017)

\bibitem{lv2021progressive}
Lv, F., Chen, X., Huang, Y., Duan, L., Lin, G.: Progressive modality
  reinforcement for human multimodal emotion recognition from unaligned
  multimodal sequences. In: Proceedings of the IEEE/CVF Conference on Computer
  Vision and Pattern Recognition. pp. 2554--2562 (2021)

\bibitem{mercea2022temporal}
Mercea, O.B., Hummel, T., Koepke, A.S., Akata, Z.: Temporal and cross-modal
  attention for audio-visual zero-shot learning. In: European Conference on
  Computer Vision. pp. 488--505. Springer (2022)

\bibitem{mercea2022audio}
Mercea, O.B., Riesch, L., Koepke, A., Akata, Z.: Audio-visual generalised
  zero-shot learning with cross-modal attention and language. In: Proceedings
  of the IEEE/CVF conference on computer vision and pattern recognition. pp.
  10553--10563 (2022)

\bibitem{miech2017learnable}
Miech, A., Laptev, I., Sivic, J.: Learnable pooling with context gating for
  video classification. arXiv preprint arXiv:1706.06905  (2017)

\bibitem{morgado2020learning}
Morgado, P., Li, Y., Nvasconcelos, N.: Learning representations from
  audio-visual spatial alignment. Advances in Neural Information Processing
  Systems  \textbf{33},  4733--4744 (2020)

\bibitem{morgado2021audio}
Morgado, P., Vasconcelos, N., Misra, I.: Audio-visual instance discrimination
  with cross-modal agreement. In: Proceedings of the IEEE/CVF conference on
  computer vision and pattern recognition. pp. 12475--12486 (2021)

\bibitem{nadeem2023sem}
Nadeem, A., Hilton, A., Dawes, R., Thomas, G., Mustafa, A.: Sem-pos:
  Grammatically and semantically correct video captioning. In: Proceedings of
  the IEEE/CVF Conference on Computer Vision and Pattern Recognition. pp.
  2605--2615 (2023)

\bibitem{nadeem2024cad}
Nadeem, A., Hilton, A., Dawes, R., Thomas, G., Mustafa, A.: Cad-contextual
  multi-modal alignment for dynamic avqa. In: Proceedings of the IEEE/CVF
  Winter Conference on Applications of Computer Vision. pp. 7251--7263 (2024)

\bibitem{nadeem2024narrativebridge}
Nadeem, A., Sardari, F., Dawes, R., Husain, S.S., Hilton, A., Mustafa, A.:
  Narrativebridge: Enhancing video captioning with causal-temporal narrative.
  arXiv preprint arXiv:2406.06499  (2024)

\bibitem{nagrani2021attention}
Nagrani, A., Yang, S., Arnab, A., Jansen, A., Schmid, C., Sun, C.: Attention
  bottlenecks for multimodal fusion. Advances in neural information processing
  systems  \textbf{34},  14200--14213 (2021)

\bibitem{okazaki2021multi}
Okazaki, S., Kong, Q., Yoshinaga, T.: A multi-modal fusion approach for
  audio-visual scene classification enhanced by clip variants. In: DCASE. pp.
  95--99 (2021)

\bibitem{ong2018deep}
Ong, E.J., Husain, S.S., Bober-Irizar, M., Bober, M.: Deep architectures and
  ensembles for semantic video classification. IEEE Transactions on Circuits
  and Systems for Video Technology  \textbf{29}(12),  3568--3582 (2018)

\bibitem{planamente2022domain}
Planamente, M., Plizzari, C., Alberti, E., Caputo, B.: Domain generalization
  through audio-visual relative norm alignment in first person action
  recognition. In: Proceedings of the IEEE/CVF winter conference on
  applications of computer vision. pp. 1807--1818 (2022)

\bibitem{poppe2010survey}
Poppe, R.: A survey on vision-based human action recognition. Image and vision
  computing  \textbf{28}(6),  976--990 (2010)

\bibitem{powers2020evaluation}
Powers, D.M.: Evaluation: from precision, recall and f-measure to roc,
  informedness, markedness and correlation. arXiv preprint arXiv:2010.16061
  (2020)

\bibitem{praveen2023recursive}
Praveen, R.G., Granger, E., Cardinal, P.: Recursive joint attention for
  audio-visual fusion in regression based emotion recognition. In: ICASSP
  2023-2023 IEEE International Conference on Acoustics, Speech and Signal
  Processing (ICASSP). pp.~1--5. IEEE (2023)

\bibitem{praveen2022joint}
Praveen, R.G., de~Melo, W.C., Ullah, N., Aslam, H., Zeeshan, O., Denorme, T.,
  Pedersoli, M., Koerich, A.L., Bacon, S., Cardinal, P., et~al.: A joint
  cross-attention model for audio-visual fusion in dimensional emotion
  recognition. In: Proceedings of the IEEE/CVF conference on computer vision
  and pattern recognition. pp. 2486--2495 (2022)

\bibitem{reddy1976speech}
Reddy, D.R.: Speech recognition by machine: A review. Proceedings of the IEEE
  \textbf{64}(4),  501--531 (1976)

\bibitem{rehman2021deep}
Rehman, A., Belhaouari, S.B.: Deep learning for video classification: A review
  (2021)

\bibitem{shah2022audio}
Shah, A., Geng, S., Gao, P., Cherian, A., Hori, T., Marks, T.K., Le~Roux, J.,
  Hori, C.: Audio-visual scene-aware dialog and reasoning using audio-visual
  transformers with joint student-teacher learning. In: ICASSP 2022-2022 IEEE
  International Conference on Acoustics, Speech and Signal Processing (ICASSP).
  pp. 7732--7736. IEEE (2022)

\bibitem{sharafi2023audio}
Sharafi, M., Yazdchi, M., Rasti, J.: Audio-visual emotion recognition using
  k-means clustering and spatio-temporal cnn. In: 2023 6th International
  Conference on Pattern Recognition and Image Analysis (IPRIA). pp.~1--6. IEEE
  (2023)

\bibitem{simonyan2014two}
Simonyan, K., Zisserman, A.: Two-stream convolutional networks for action
  recognition in videos. Advances in neural information processing systems
  \textbf{27} (2014)

\bibitem{song2022multimodal}
Song, Q., Sun, B., Li, S.: Multimodal sparse transformer network for
  audio-visual speech recognition. IEEE Transactions on Neural Networks and
  Learning Systems  (2022)

\bibitem{soomro2012ucf101}
Soomro, K., Zamir, A.R., Shah, M.: Ucf101: A dataset of 101 human actions
  classes from videos in the wild. arXiv preprint arXiv:1212.0402  (2012)

\bibitem{srivastava2014dropout}
Srivastava, N., Hinton, G., Krizhevsky, A., Sutskever, I., Salakhutdinov, R.:
  Dropout: a simple way to prevent neural networks from overfitting. The
  journal of machine learning research  \textbf{15}(1),  1929--1958 (2014)

\bibitem{sun2022human}
Sun, Z., Ke, Q., Rahmani, H., Bennamoun, M., Wang, G., Liu, J.: Human action
  recognition from various data modalities: A review. IEEE transactions on
  pattern analysis and machine intelligence  (2022)

\bibitem{sundar2022multimodal}
Sundar, A., Heck, L.: Multimodal conversational ai: A survey of datasets and
  approaches. arXiv preprint arXiv:2205.06907  (2022)

\bibitem{tsai2019multimodal}
Tsai, Y.H.H., Bai, S., Liang, P.P., Kolter, J.Z., Morency, L.P., Salakhutdinov,
  R.: Multimodal transformer for unaligned multimodal language sequences. In:
  Proceedings of the conference. Association for computational linguistics.
  Meeting. vol.~2019, p.~6558. NIH Public Access (2019)

\bibitem{tschannen2020self}
Tschannen, M., Djolonga, J., Ritter, M., Mahendran, A., Houlsby, N., Gelly, S.,
  Lucic, M.: Self-supervised learning of video-induced visual invariances. In:
  Proceedings of the IEEE/CVF Conference on Computer Vision and Pattern
  Recognition. pp. 13806--13815 (2020)

\bibitem{vaswani2017attention}
Vaswani, A., Shazeer, N., Parmar, N., Uszkoreit, J., Jones, L., Gomez, A.N.,
  Kaiser, {\L}., Polosukhin, I.: Attention is all you need. Advances in neural
  information processing systems  \textbf{30} (2017)

\bibitem{vilaça2022recent}
Vilaça, L., Yu, Y., Viana, P.: Recent advances and challenges in deep
  audio-visual correlation learning (2022)

\bibitem{wang2021semantic}
Wang, G., Chen, C., Fan, D.P., Hao, A., Qin, H.: From semantic categories to
  fixations: A novel weakly-supervised visual-auditory saliency detection
  approach. In: Proceedings of the IEEE/CVF conference on computer vision and
  pattern recognition. pp. 15119--15128 (2021)

\bibitem{wang2017monkeytyping}
Wang, H.D., Zhang, T., Wu, J.: The monkeytyping solution to the youtube-8m
  video understanding challenge. arXiv preprint arXiv:1706.05150  (2017)

\bibitem{wei2022learning}
Wei, Y., Hu, D., Tian, Y., Li, X.: Learning in audio-visual context: A review,
  analysis, and new perspective (2022)

\bibitem{Wu_2019_ICCV}
Wu, Y., Zhu, L., Yan, Y., Yang, Y.: Dual attention matching for audio-visual
  event localization. In: Proceedings of the IEEE/CVF International Conference
  on Computer Vision (ICCV) (October 2019)

\bibitem{yu2022mm}
Yu, J., Cheng, Y., Zhao, R.W., Feng, R., Zhang, Y.: Mm-pyramid: Multimodal
  pyramid attentional network for audio-visual event localization and video
  parsing. In: Proceedings of the 30th ACM International Conference on
  Multimedia. pp. 6241--6249 (2022)

\bibitem{zhang2023variational}
Zhang, J., Yu, Y., Tang, S., Wu, J., Li, W.: Variational autoencoder with cca
  for audio--visual cross-modal retrieval. ACM Transactions on Multimedia
  Computing, Communications and Applications  \textbf{19}(3s),  1--21 (2023)

\bibitem{zhang2017learning}
Zhang, S., Zhang, S., Huang, T., Gao, W., Tian, Q.: Learning affective features
  with a hybrid deep model for audio--visual emotion recognition. IEEE
  Transactions on Circuits and Systems for Video Technology  \textbf{28}(10),
  3030--3043 (2017)

\bibitem{zhang2023facial}
Zhang, Z., An, L., Cui, Z., Dong, T., et~al.: Facial affect recognition based
  on transformer encoder and audiovisual fusion for the abaw5 challenge. arXiv
  preprint arXiv:2303.09158  (2023)

\bibitem{zhu2020deep}
Zhu, H., Luo, M., Wang, R., Zheng, A., He, R.: Deep audio-visual learning: A
  survey (2020)

\end{thebibliography}
\end{document}